\documentclass[11pt]{article}

\usepackage[final]{acl}

\usepackage{times}
\usepackage{latexsym}
\usepackage{listings}

\usepackage[T1]{fontenc}
\usepackage[utf8]{inputenc}

\usepackage{microtype}

\usepackage{inconsolata}

\usepackage{graphicx}

\title{Science Consultant Agent}

\author{
    Karthikeyan K$^{1}$\thanks{Work done as an intern at Amazon},
    Philip Wu$^{2}$,
    Xin Tang$^{2}$, Alexandre Alves $^{2}$\\
    $^{1}$Department of Computer Science, Duke University
    $^{2}$Amazon\\
    \texttt{karthikeyan.k@duke.edu}
    \texttt{\{phil,  xintang, alvesa\}@amazon.com}
}
\begin{document}
\maketitle

\begin{abstract}

The Science Consultant Agent is a web-based Artificial Intelligence (AI) tool that helps practitioners select and implement the most effective modeling strategy for AI-based solutions. It operates through four core components: Questionnaire, Smart Fill, Research-Guided Recommendation, and Prototype Builder. By combining structured questionnaires, literature-backed solution recommendations, and prototype generation, the Science Consultant Agent accelerates development for everyone from Product Managers and Software Developers to Researchers. The full pipeline is illustrated in Figure~\ref{fig:pipeline}.

\end{abstract}

\section{Introduction}

AI practitioners—including applied scientists, engineers, and product managers—face a critical challenge in selecting the optimal modeling strategy for a given task. The decision space spanning prompting frontier large language models (LLMs)~\citep{wei2022chain}, implementing Retrieval-Augmented Generation (RAG)~\citep{lewis2020retrieval}, fine-tuning domain-specific models~\cite{hu2021loralowrankadaptationlarge}, distilling knowledge from larger models~\cite{hinton2015distillingknowledgeneuralnetwork}, and developing other specialized techniques is complex, rapidly evolving, and highly contextual. Each strategy has distinct advantages and limitations, with specific requirements and implications that make the decision difficult.

Without structured guidance, practitioners, especially non-research users, often default to seemingly accessible solutions such as direct prompting or RAG~\cite{Luchins1942MechanizationIP}. This tendency is reinforced by \emph{example-induced bias}, a phenomenon where teams design prompts or instructions around familiar examples, inadvertently shaping their entire approach around these narrow cases~\cite{doi:10.1126/science.185.4157.1124}. When the model successfully handles such examples, it creates a misleading perception that the broader task has been solved. In practice, this reflects overfitting to specific instructions rather than robust generalization across diverse, real-world scenarios. As a result, teams systematically misallocate resources: over-investing in costly LLM-based methods when smaller, specialized models would suffice, or applying generic approaches to tasks that require domain-specific capabilities.

\begin{figure}[!htbp]
  \centering
  \includegraphics[width=0.99\linewidth]{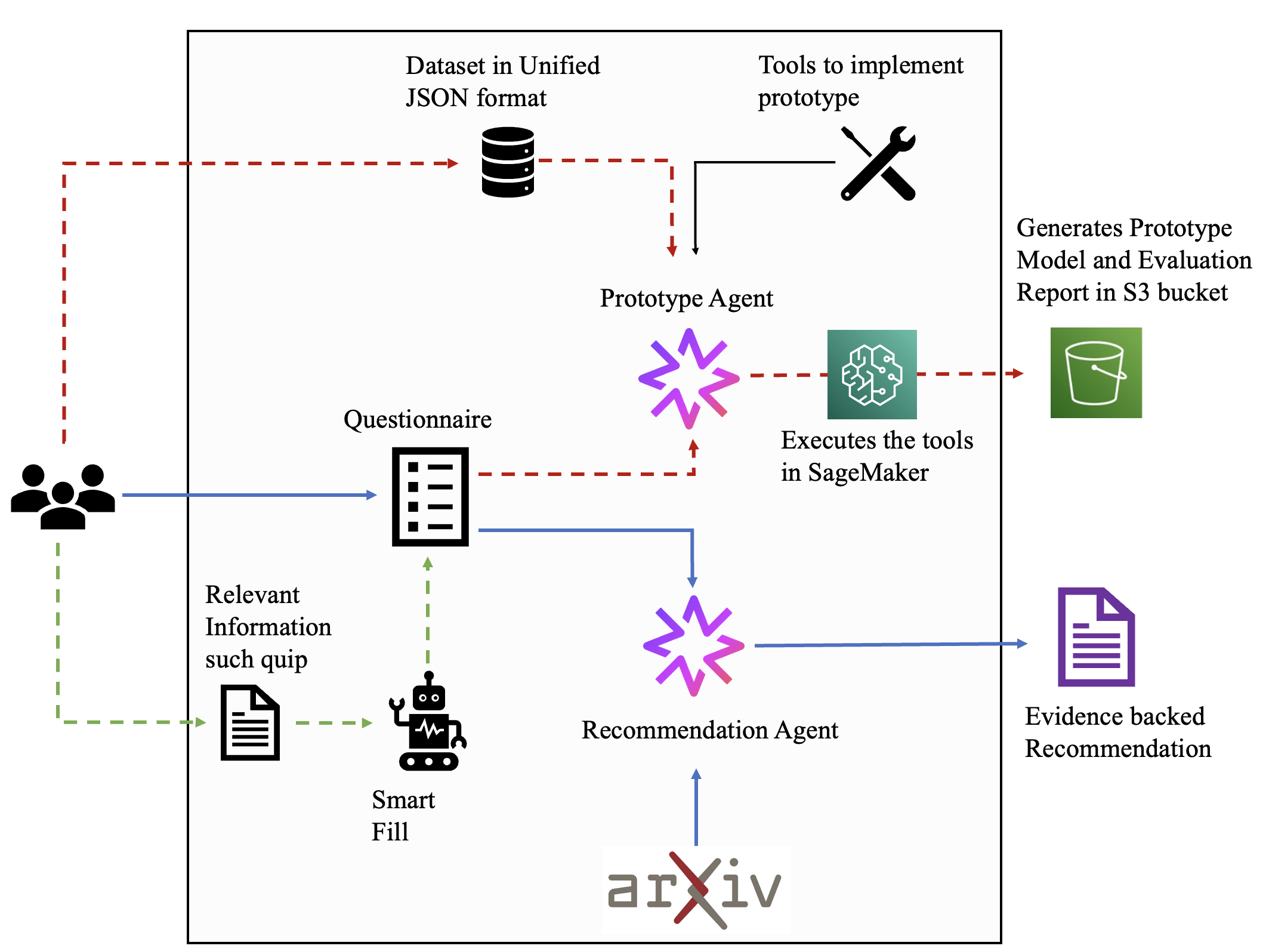}
  \caption{The full pipeline of the Science Agent.}
  \label{fig:pipeline}
\end{figure}

These inefficiencies are amplified in the current LLM era, where modeling strategies incur substantial computational costs~\citep{cottier2025risingcoststrainingfrontier}. Brute-force exploration across many options, as in traditional AutoML~\citep{he2021automl}, is time-consuming and resource-intensive. For LLM-based methods, even limited experimentation with prompt variants or multiple fine-tuning runs can quickly become prohibitively expensive. In such settings, exhaustive search is neither practical nor sustainable, making disciplined and evidence-driven decisions essential.

To address these challenges, we present the \emph{Science Consultant Agent}, a web-based AI agent designed to guide practitioners toward disciplined, evidence-based modeling decisions. The Agent consists of four components: (1) \emph{Questionnaire}, which ensures task requirements, data characteristics, and constraints are systematically captured; (2) \emph{Smart Fill}, which leverages project descriptions and metadata to auto-complete many of these fields, reducing user effort; (3) \emph{Research-Guided Recommendation}, which searches arXiv and generates literature-backed recommendations; and (4) \emph{Prototype Builder}, which takes user-provided datasets and automatically implements standard baselines using Amazon SageMaker. Together, these components form a unified workflow that serves diverse users. For product managers, the Agent enables rapid prototyping and encourages early consideration of requirements and trade-offs. For engineers and developers, it provides research-backed strategies that reduce wasted effort and support higher-quality technical decisions. For scientists, it acts as a literature survey assistant, streamlining the discovery of relevant work and emerging research directions.

Evaluation of the Science Agent is challenging, particularly because judging the best modeling strategy would require implementing and tuning many alternatives, which is not feasible. To address this, we rely on user feedback for evaluation. The feedback showed that the recommendations often matched users’ expectations and that the justifications were convincing, but it also revealed that some terminologies in questionnaire was unclear and that guidance for transforming research recommendations into implementations was limited.

The contribution of this work is an end-to-end agent that integrates structured guidance, literature-grounded recommendations, and prototype generation into a single workflow. By connecting task specification, evidence-based recommendation, and automated prototyping, the Science Consultant Agent makes disciplined modeling decisions accessible and practical for a wide range of users.

\section{Science Consultant Agent Overview}

In this section, we describe each component of the Science Agent in detail.

\subsection{Questionnaire}

The Science Agent begins with a six-part questionnaire designed to fully capture the user’s task. The six parts are:
\begin{enumerate}
\item \textbf{Introduction}: This part contains questions asking for a brief description of the task, the business problem to be solved, relevant key performance indicators (KPIs), etc.  
\item \textbf{Understanding Data}: This part contains questions about the domain, quality, and availability of the training, validation, and test data.  
\item \textbf{Evaluation}: This part contains questions about evaluation and metrics.  
\item \textbf{Task Mechanism}: This part contains questions regarding the capabilities—such as real-time information, API/tool access, and specific reasoning abilities—required for the task.  
\item \textbf{Constraints}: This part contains questions regarding constraints such as latency, cost, and performance trade-offs.  
\item \textbf{Miscellaneous}: This part includes other miscellaneous questions such as the need for interpretability, whether the model needs to be updated frequently, and any existing baselines.  
\end{enumerate}
We ask the user to fill out this questionnaire, and the responses are used to understand the task and make the recommendation. \\

\noindent \textbf{Why Questionnaire:}  
Many users do not know which questions to ask or what information an LLM needs to understand the task. The Science Agent’s pre-set, structured questions systematically gather this information, rather than relying on the user to guess. Additionally, the questionnaire has an educational role: it encourages users to think about their projects clearly and in a structured manner. It also ensures that key trade-offs and design considerations—such as latency, cost, performance, and evaluation metrics—are addressed early.

\noindent \textbf{Feedback and Future Improvements.}  
Based on surveys and interviews, we received feedback that the questionnaire contained too many questions and that some terminology was unclear, particularly for non-scientists. We also found that many questions were left unanswered or answered incorrectly. Motivated by this feedback, we integrated Smart Fill to auto-complete many fields from the project description, reducing the work required from users. We also plan to introduce role-tailored questionnaires: scientists will receive a minimal set of essential technical questions, while non-scientists will focus on higher-level project descriptions, with the LLM generating follow-up questions only when needed.

\subsection{Smart Fill}

Motivated by the feedback on questionnaire length, we integrated Smart Fill to auto-complete many fields using the introduction questions or other project documents. Smart Fill uses its internal knowledge and contextual reasoning to answer most of the questionnaire. For example, based on the project description, Smart Fill can suggest appropriate evaluation metrics—such as when to choose precision, accuracy, or AUC-ROC—identify whether latency or high performance is likely to be important, or infer which reasoning capabilities the task may require.  

In contrast, questions about data characteristics—especially data availability—cannot be answered reliably from internal knowledge alone. Determining data availability is challenging not only for LLMs but also for humans, since it often requires reviewing existing datasets and assessing their relevance. To address this, Smart Fill combines the project description with metadata from internal tables, allowing the LLM to identify datasets that may be relevant to the task; Smart Fill then uses this information to answer the data availability questions.  

Overall, Smart Fill helps reduce the effort required to complete the questionnaire: instead of answering all the questions manually, users receive auto-completed responses that they can review and edit as needed.

\begin{figure*}[!htbp]
  \centering
  \includegraphics[width=0.8\linewidth]{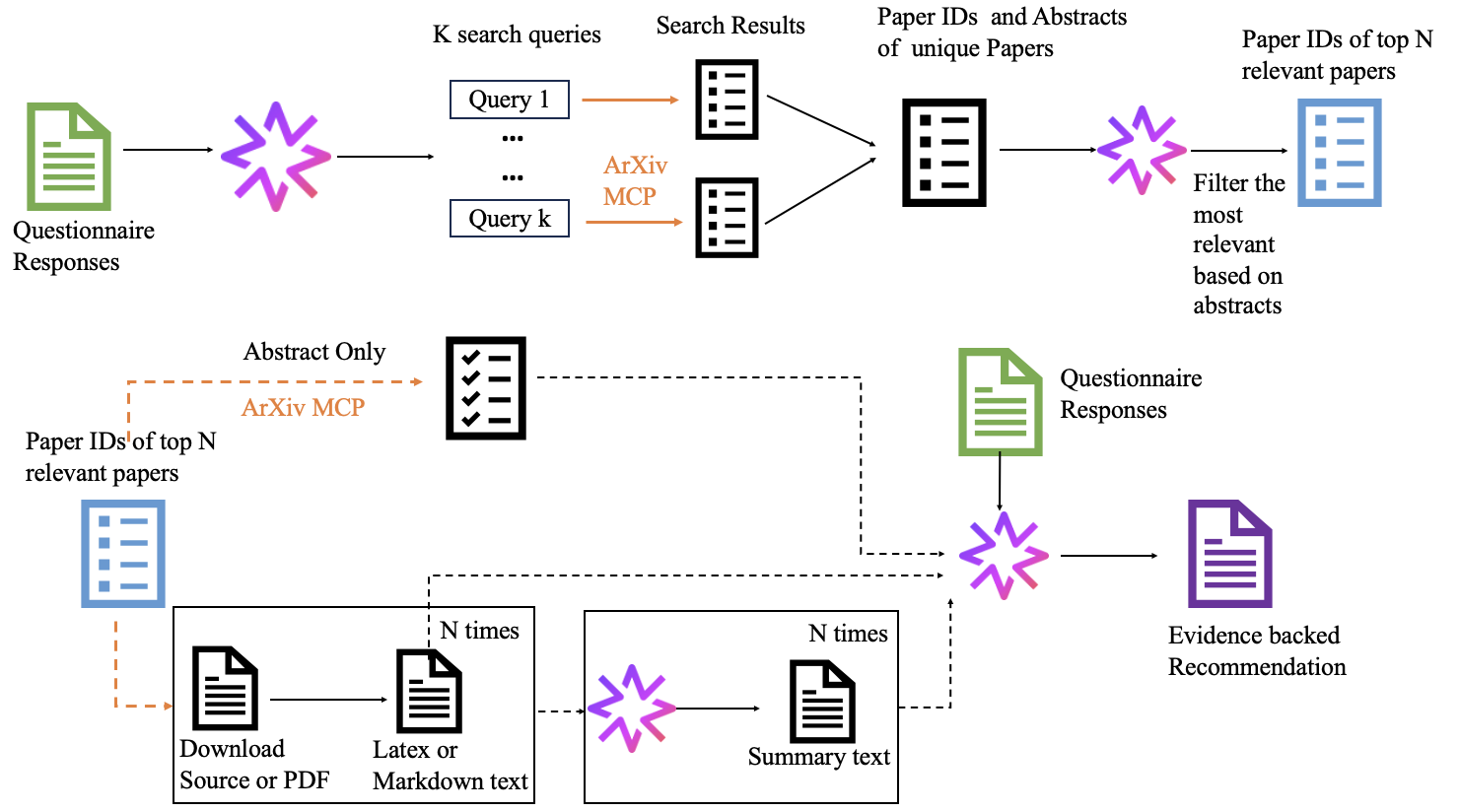}
  \caption{Evidence Based Recommendation Generation.}
  \label{fig:evidence}
\end{figure*}

\subsection{Evidence-Based Recommendation}

To generate evidence-based recommendations, we first retrieve relevant literature from arXiv and use it as context to generate the recommendations for the most effective solution strategy.

\subsubsection{Evidence Retrieval} 
\quad \\
\noindent \textbf{Retrieval Backend: arXiv MCP}
We use an arXiv Model Context Protocol (MCP), a wrapper around the arXiv API, which provides the following functionality:
\begin{itemize}
\item \textbf{Search:} given a search query, it runs a black-box internal search and returns metadata for relevant papers (paper ID, title, abstract).
\item \textbf{Download source and PDF:} downloads the LaTeX source and PDF.
\end{itemize}

We started from the existing \texttt{arxiv-mcp-server}~\cite{arxiv-mcp-server}, which supports search, PDF download, and PDF-to-Markdown conversion using \texttt{pymupdf4llm}. However, this PDF-to-Markdown conversion fails frequently. Therefore, we adapted it to directly use LaTeX files when available. We download the LaTeX source, concatenate all \texttt{.tex} files, and pass the concatenated LaTeX file to LLMs. The major advantages of using LaTeX instead of converting PDF to Markdown are that tables and equations are preserved exactly, and that processing is much faster as there is no additional conversion step. The trade-offs are that document order can be lost as we do not enforce order when concatenating \texttt{.tex} files, and the context may include extra tokens corresponding to imports and comments. When LaTeX source is unavailable, we fall back to PDF-to-Markdown. \\

\noindent \textbf{Query Generation}  
The arXiv search interface is a black box; we cannot control its internal ranking or directly pass the questionnaire responses. To work within these constraints, we prompt the LLM to analyze the questionnaire responses and generate up to K queries (K is a hyperparameter, currently set to 50). The LLM crafts these queries following the best practices published on the arXiv website. We enforce JSON-structured output so the queries can be parsed reliably. \\

\noindent \textbf{Search and Deduplication}  
For each generated query, we use the arXiv MCP search functionality to retrieve relevant paper IDs, titles, and abstracts. We remove duplicates across queries to ensure that all retrieved papers are unique. \\

\noindent \textbf{Filtering Relevant Papers}  
We consolidate all retrieved abstracts and prompt the LLM to select up to N of the most relevant papers (N is a hyperparameter, currently set to 50) based on the abstracts. This step also uses structured output so that the resulting list of paper IDs can be directly consumed by downstream components.

\subsubsection{Context Construction}
Once the most relevant papers are identified, we construct the context provided to the LLM for final recommendation synthesis. We support three strategies to construct the context, 

\noindent \textbf{Abstract Only}  
In this strategy, we use the paper’s title, arXiv ID (with link), and abstract. This approach is extremely fast, has minimal processing overhead, and allows many diverse papers to be included in context. However, the LLM does not gain a deep understanding of the paper and often relies on prior knowledge.

\noindent \textbf{Full Paper}  
For the PDF-based variant, we download the PDF and include it as base64-encoded content (supported natively via \texttt{boto3}). Due to token limitations, only one paper can be included this way. For the text-based variant, we use extracted LaTeX or Markdown text as context, which allows one or two papers before hitting token limits. This strategy enables the LLM to gain a deep understanding of the paper. The drawback is that only one or two papers can be included, and the final recommendation is heavily biased toward these papers.

\noindent \textbf{Summaries}  
For this strategy, we combine the questionnaire responses with each paper’s PDF and prompt the LLM to produce a one-page, task-specific summary. These summaries from multiple papers are then aggregated as the context. This preserves more content than abstracts, supports multiple papers, and captures paper-specific relevance. The limitation is that it is very slow due to the computational overhead of summarization.

\subsubsection{Final Recommendation Generation}
Finally, we pass the complete questionnaire responses and the generated context to the LLM and instruct it to produce the final recommendations: the best solution and the baseline.

\subsubsection{Feedback and Future Improvements}

\noindent \textbf{Internal and External Sources}  
Reviewers suggested expanding the literature search to include internal research papers and domain-specific resources such as pharmacy or medical journals. When such resources are available, it is straightforward to integrate the resources into the system.

\noindent \textbf{Filtering Credible Sources}  
We received feedback to filter the results from arXiv so that only credible sources are used. We plan to apply filters based on citation counts and a known list of conferences and journals to ensure that our recommendations are based on credible, highly cited research papers.

\noindent \textbf{Locally Hosted and Pre-Processed Papers}  
An alternative to arXiv MCP is to host papers locally. This approach has two main advantages. First, it would enable flexible search, such as using embedding-based or other advanced retrieval strategies. Second, it would reduce latency if we pre-process papers in advance, especially for summary-based context where we could include a generic offline summary.

\subsection{Prototype-Builder}

Science Agent goes beyond recommending the best solution—it can also build a prototype, from a limited set of modeling options, and generate an evaluation report, provided the necessary data is available from the user. This capability is particularly useful for non-scientist users. We implemented a tool-based approach for prototype generation: the LLM selects from a predefined set of available tools, chooses appropriate values for any required parameters, and the chosen tool carries out the actual implementation. While this approach supports only a limited set of baselines, it is safe to execute, avoids the risks of arbitrary or malicious code execution, and ensures that the logic is always correct and that results and evaluations are reproducible and reliable.

We also considered an alternative approach, where the LLM generates Python code, executes and debugs the code. This offers greater flexibility but it also introduces serious risks: the generated code may produce misleading results or corrupt data, which can be extremely dangerous. For these reasons, we do not adopt this method at the current stage, though it may be reconsidered if LLMs become sufficiently reliable for autonomous coding.

\subsubsection{Prototype Builder Tools}

\begin{figure*}[!htbp]
  \centering
  \includegraphics[width=0.9\linewidth]{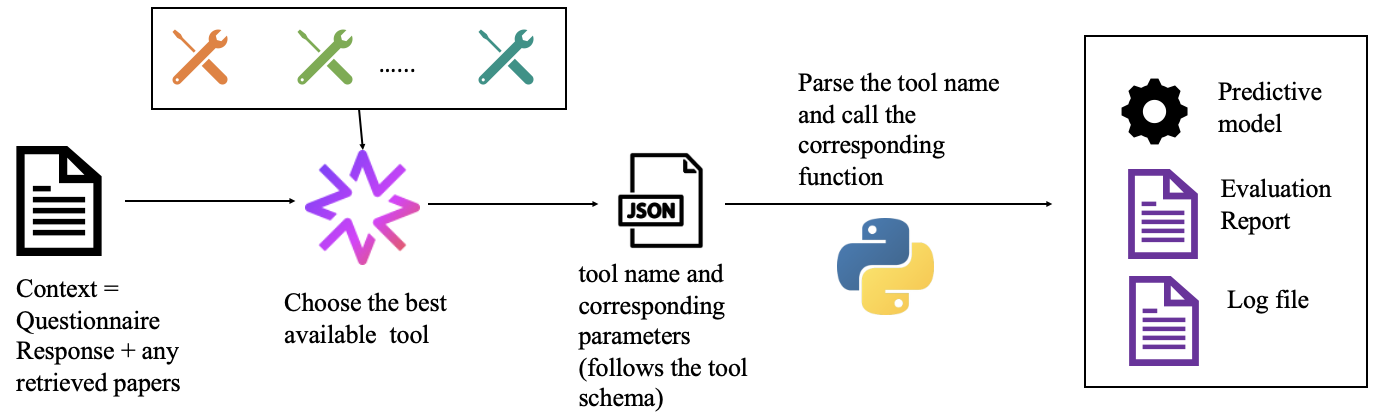}
  \caption{Prototype-Builder.}
  \label{fig:prototype}
\end{figure*}

In our tool-based implementation, each tool corresponds to a standard modeling strategy. Once the LLM understands the user task, it selects a tool, provides the required parameters, and the tool carries out the actual implementation. 

All tools are implemented as functions that accept input and output S3 paths, hyperparameters, metric names, and the SageMaker instance type. Each tool reads the data from the input S3 path, assumes it follows the \emph{Unified Data Template Structure}, and validates that the format is correct for the intended modeling approach. For example, the XGBoost prototype builder requires features and labels to be numerical or categorical, consistent across examples. Tools also validate hyperparameters and check that the specified S3 paths are accessible. After validation, the tool submits a training job through SageMaker, evaluates the model on the test set, and writes all artifacts—including the trained model, predictions, evaluation metrics, and logs—to the output S3 path. \\

\noindent\textbf{Tabular Data Tools: } Our tools for tabular data support supervised tasks, specifically regression, binary classification, and multi-class classification. We use \texttt{autogluon.cloud}'s \texttt{TabularCloudPredictor}~\cite{erickson2020autogluontabularrobustaccurateautoml} to submit training jobs on SageMaker and manage model artifacts. We support various modeling strategies, including gradient boosting methods (XGBoost, LightGBM, CatBoost), neural networks (FastAI and Torch), tabular transformers, and ensemble methods such as bagging, stacking, and distillation. We also support various metrics for validation and final reporting, depending on the problem type. \\

\noindent\textbf{Text Data Tools: } For text data, our current tools include prompting baselines, specifically direct and chain-of-thought prompting. These tools can generate predictions but do not yet produce evaluation reports, since evaluation for text generation is not straightforward. The difficulty arises because model outputs and ground truth answers often differ in surface form even when they are correct. For example, if the ground truth answer is “8400,” the model might output “8,400,” “\$8400,” or “8400 USD.” Similarly, if the answer is “45,” the model might output “45 mph” or “45 miles per hour.” Although semantically correct, such differences cause simple string matching to fail. These nuances make evaluation for text data considerably more difficult than for tabular tasks.

\subsubsection{Future Improvements}

While the current tools establish the feasibility of prototype building, we could further extend them to support a wider range of tasks—such as clustering, fine-tuning, and distillation—and data modalities, including images and audio, for broader applicability.  Evaluation for generative tasks remains a key challenge. Beyond string matching, we could consider LLM-as-judge, in which an LLM determines correctness by comparing predictions to ground truth. Although more flexible, this approach adds compute cost and may also produce incorrect evaluations.  A more ambitious improvement would be to build an agent that can read a research paper and its associated GitHub repository, interpret the implementation details, and replicate the code on our dataset. This approach is powerful but risky: LLM-generated code can produce misleading results or even corrupt data, which can be extremely dangerous. Such functionality should only be used under expert supervision and never with sensitive datasets.

\section{Evaluation}

While the Science Agent has four components, recommendation generation is the primary component that requires thorough evaluation; the questionnaire is manually written, Smart Fill mainly assists the user (who will review and edit), and the Prototype-Builder uses tools written by us. However, evaluating recommendation quality is challenging: ideally we would implement alternative modeling strategies and compare whether the recommended solution yields the best results. Furthermore, as research advances continually, the target for “best” shifts, making stable benchmarks difficult. Given these constraints, we rely on user feedback through surveys and one-on-one follow-up interviews.

\subsection{User Study and Findings}

The Science Agent website includes a survey at the end, and we use the survey responses to evaluate the system. The survey asks users about their familiarity with AI research, whether the recommendations align with their expectations, whether the justifications are convincing, and their overall experience. In addition to the survey, we conduct one-on-one interviews, particularly with non-research users, to gather deeper insights. We organize the testing in two rounds.  

\noindent\textbf{Round 1:} In the first round, the audience mainly includes the internal applied science team. At this stage, recommendations are generated solely from the LLM’s internal knowledge, without any grounding in arXiv or other external sources. Based on the survey results, 82\% of participants rate the overall experience as excellent or good; 100\% find the justifications convincing (with 73\% rating them moderately to very convincing); and 100\% observe alignment with their expectations (with 63\% rating the alignment as moderate to perfect).

\noindent\textbf{Round 2.} We conducted the second round of testing after grounding the recommendations in arXiv papers. This round focused on non-researchers, specifically seven PMs and SDEs, with one-on-one follow-ups for four of them. One user tested the system on a project they were already working on and reported that the recommendations aligned with their expectations but did not provide many new insights.  Several users showed strong interest in Smart-Fill, especially data discovery, as well as in the Prototype-Builder. 
%They noted that data discovery is a very challenging task even for human experts and that support from an LLM agent would be highly desirable. 
One user mentioned being unsure how to use the recommendations and said it would help if the agent assisted further, for example by identifying relevant contacts, estimating task complexity, estimating required computational resources, or providing insights that directly answer specific analysis questions. Finally, a few users noted that certain terminology in the questionnaire was unclear and required them to look it up.

\subsection{Next Steps in Evaluation}

We plan to collect higher-quality questionnaire responses and have human experts assess whether recommendations are meaningful, aligned with their expectations, and supported by convincing justifications. Since this may not scale and individual biases can affect scoring, we also consider comparative formats where reviewers rank two recommendation variants for the same questionnaire response. When human evaluation is impractical, another LLM could serve as a judge to determine which recommendation appears more compelling.

\section{Conclusion}

We built Science Consultant Agent, an end-to-end workflow with four components: Questionnaire, Smart Fill, Evidence-Based Recommendation, and Prototype-Builder. We described the challenges encountered, the feedback received, and directions for improving each component. Together, these contributions address the difficulty of selecting modeling strategies in a rapidly evolving landscape and make evidence-based decisions more accessible.

\section{Limitations}

The current version of the Science Consultant Agent is an initial prototype, and each of its four components can be further optimized to be more effective. In addition, evaluation remains challenging and limited, as it primarily relies on user feedback; future work should explore more thorough and systematic evaluation methods.

\bibliography{custom}

\appendix
\section{Appendix}

\section*{Appendix: Unified Data Template}

Each of the training, validation and test sata must be provided in \textbf{JSONL} format, one JSON object per line. 
Each json object has the following fixed top-level keys:

\begin{itemize}
    \item \textbf{unique\_id} (string) -- Unique identifier.
    \item \textbf{input} (object, optional):
    \begin{itemize}
        \item \texttt{text} (string)
        \item \texttt{image\_url}, \texttt{audio\_url}, \texttt{video\_url} (S3 URIs)
        \item \texttt{base64} (string, binary data)
        \item \texttt{numerical\_features} (map: key $\rightarrow$ number)
        \item \texttt{categorical\_features} (map: key $\rightarrow$ string)
    \end{itemize}
    \item \textbf{output} (object, optional):
    \begin{itemize}
        \item \texttt{text}, \texttt{numerical}, \texttt{categorical}
        \item \texttt{character\_spans}: \{``start\_char'', ``end\_char''\}
    \end{itemize}
\end{itemize}

\subsection*{Principles}
\begin{enumerate}
    \item Fields are optional; task-specific tools enforce required keys.
    \item Top-level keys are fixed globally; user-defined keys appear only inside feature maps.
    \item Binary data is referenced via S3 URIs or Base64 encoding.
\end{enumerate}

\section{Recommendation Generation Prompt}

\begin{quote}
Analyze the following questionnaire response and the provided summaries of relevant research papers. 
Your goal is to recommend the best approach to solve the user task. You will provide two separate 
recommendations: Best Solution and Strong Baseline, each with a clear description, justification, 
and references.

\medskip

Begin each response with a thinking phase inside \texttt{<small><em>...</em></small>} tags. In this 
phase, think about what the best solution and strong baselines for the task might be, which citations 
or supporting evidence you plan to use, and how you will justify your recommendations. Clearly 
identify and list the citations you intend to reference, using the appropriate citation format. Inside 
the thinking use simple paragraph, do not use markdown format.

\medskip

1. Best Approach to Solve the Task:  
\hspace{1em}-- This recommendation should present the best solution for the task and is likely to 
achieve state-of-the-art results. You must always provide justification with relevant citations from 
recent research papers published in reputable conferences or journals.

\medskip

2. Strong Baseline:  
\hspace{1em}-- This recommendation should suggest a strong and widely recognized baseline. Again, 
always provide justification with relevant citations or logical arguments explaining why this is a 
strong baseline. Common examples of baselines include, but are not limited to: gradient boosting 
methods (e.g., XGBoost), random forest, simple prompting, chain-of-thought prompting, retrieval-
augmented generation (RAG), knowledge distillation, diffusion-based generative approaches, fine-tuning 
with contrastive learning, reinforcement learning (e.g., PPO, SAC, DQN), time series forecasting with 
modern boosting or neural approaches, or any other well-established and competitive approach relevant 
to the task.

\medskip

-------------------------  
\textbf{Questionnaire Response:}  
\texttt{\{formatted\_qa\}}  
-------------------------  
\textbf{Summaries of Relevant Papers:}  
\texttt{\{summaries\_str\}}  
-------------------------  

Your response should follow the Markdown format below for each of the two recommendations.

\medskip

\#\# Best Solution

\#\# Strong Baseline

Within each recommendation, include the following sections:

\medskip

\textbf{Description}  
A brief one or two paragraph description of the recommended solution.

\medskip

\textbf{Step-by-Step Solution}  
Detail the solution clearly enough for a Machine Learning or AI Engineer or SDE to implement. Ideally 
even a Product Manager should understand the solution and communicate clearly to an SDE to implement. 
Typical details often include but not limited to:  
-- \textbf{Data:} Details regarding the data, such as what data is used, any required preprocessing 
steps, and the expected inputs and outputs, etc. Maybe even show an example if possible  
-- \textbf{Modeling:} Details regarding the modeling approach, such as LLMs or model architectures, 
learning algorithms, objective functions, etc.  
-- \textbf{Prediction:} How predictions are made—whether they are direct outputs from the model or 
require further processing, etc.  
-- \textbf{Evaluation:} Details regarding evaluation, such as ground truth, metrics to use, etc.  
-- Any other relevant implementation details needed for clarity. The above steps are general guidelines 
based on the project some parts may not be relevant or in some cases you might have to include more 
details to be clear. 

\medskip

\textbf{Coding Details}  
Provide a brief design doc describing key components and their roles, with a concise pseudocode block 
showing only class/function headers and core control flow (no imports or full implementations). Put 
the pseudocode in a fenced Markdown code block.

\medskip

\textbf{Justification}  
A strong, evidence-based justification for why this is the most suitable recommendation, supported by 
relevant citations. Always use actual author names and years from the source when citing. Format 
citations as follows: (Author, Year) for one author; (Author \& Author, Year) for two authors; 
(Author et al., Year) for three or more authors. Never use placeholders or generic names such as 
(Author1 et al., Year). Do not hallucinate citations.

\medskip

\textbf{References}  
List all cited sources here.

\medskip
-------------------------
\end{quote}

\end{document}